\documentclass[10pt,twocolumn,letterpaper]{article}

\usepackage{cvpr}              
\usepackage[accsupp]{axessibility}
\usepackage{graphicx}
\usepackage{amsmath}
\usepackage{amssymb}
\usepackage{booktabs}
\usepackage{multirow}
\usepackage{pifont}
\usepackage[table]{xcolor}
\usepackage[normalem]{ulem}

\usepackage{dsfont}
\usepackage[pagebackref,breaklinks,colorlinks]{hyperref}
\makeatletter
\@namedef{ver@everyshi.sty}{}
\makeatother
\usepackage{pgfplots}
\pgfplotsset{compat=1.3}
\usepackage[capitalize]{cleveref}
\crefname{section}{Sec.}{Secs.}
\Crefname{section}{Section}{Sections}
\Crefname{table}{Table}{Tables}
\crefname{table}{Tab.}{Tabs.}
\usepackage{amssymb}
\usepackage{pifont}
\newcommand{\xmark}{\ding{55}}%

\def\confName{CVPR}
\def\confYear{2022}

\usepackage{xcolor}
\makeatletter
\DeclareRobustCommand{\cev}[1]{%
  {\mathpalette\do@cev{#1}}%
}
\newcommand{\do@cev}[2]{%
  \vbox{\offinterlineskip
    \sbox\z@{$\m@th#1 x$}%
    \ialign{##\cr
      \hidewidth\reflectbox{$\m@th#1\vec{}\mkern4mu$}\hidewidth\cr
      \noalign{\kern-\ht\z@}
      $\m@th#1#2$\cr
    }%
  }%
}
\makeatother

\newcommand{\bv}{\mathbf{v}}
\newcommand{\ovl}[1]{\cev{#1}}
\newcommand{\ovr}[1]{\vec{#1}}
\newcommand{\rt}{\ovr{t}}
\newcommand{\lt}{\ovl{t}}
\newcommand{\rbv}{\ovr{\bv}}
\newcommand{\lbv}{\ovl{\bv}}
\newcommand{\ru}{\ovr{u}}
\newcommand{\lu}{\ovl{u}}
\newcommand{\rv}{\ovr{v}}
\newcommand{\lv}{\ovl{v}}
\newcommand{\rw}{\ovr{w}}
\newcommand{\lw}{\ovl{w}}
\newcommand{\rM}{\ovr{M}}
\newcommand{\lM}{\ovl{M}}
\newcommand{\J}{\mathcal{J}}
\newcommand{\F}{\mathcal{F}}

\begin{document}

\title{Accelerating Video Object Segmentation with Compressed Video}

\author{
 Kai Xu \quad Angela Yao\\
 National University of Singapore \\
 {\tt\small \{kxu, ayao\}@comp.nus.edu.sg}
}
\maketitle

\begin{abstract}
We propose an efficient plug-and-play acceleration framework for semi-supervised video object segmentation by exploiting the temporal redundancies in videos presented by the compressed bitstream.
Specifically, we propose a motion vector-based warping method for propagating segmentation masks from keyframes to other frames in a bi-directional and multi-hop manner.  Additionally, we introduce a residual-based correction module that can fix wrongly propagated segmentation masks from noisy or erroneous motion vectors.  Our approach is flexible and can be added on top of several existing video object segmentation algorithms. We achieved highly competitive results on DAVIS17 and YouTube-VOS on various base models with substantial speed-ups of up to 3.5X with minor drops in accuracy. \footnote{Code: \url{https://github.com/kai422/CoVOS}}  
\end{abstract}

\section{Introduction}
\label{sec:intro}

Video object segmentation (VOS) aims to obtain pixel-level masks of the objects in a video sequence.  State-of-the-art methods~\cite{premvos,OSVOS-S,STM,masktrack} are highly accurate at segmenting the objects, but they can be slow, requiring as much as 0.2 seconds~\cite{STM} to segment a frame.  More efficient methods~\cite{FRTM,SiamMask,Track-Seg} typically trade off accuracy for speed.  

To minimize this trade-off, we propose to leverage compressed videos for accelerating video object segmentation. Most videos on the internet today are stored and transmitted in a compressed format. Video compression encoders take a sequence of raw images as input and exploit the inherent spatial and temporal redundancies to compress the size by several magnitudes~\cite{le1991mpeg}. The encoding gives several sources of ``free'' information for VOS.  Firstly, the bitstream's frame type (I- vs. P-/B-frames) gives some indication for keyframes, as the encoder separates the frames according to their information content. Secondly, the motion compensation scheme used in compression provides motion vectors that serve as a cheap approximation to optical flow. Finally, the residuals 
give a strong indicator of problematic areas that may require refinement.  

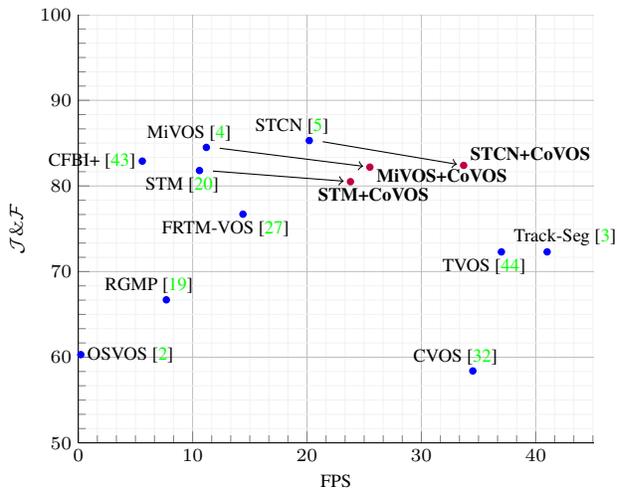
\begin{figure}[t!]
\begin{tikzpicture}
\scriptsize

\begin{axis}[xmin=0,ymin=50, ymax=100,
    axis x line*=bottom,
    axis y line*=left,
    ylabel = $\mathcal{J}\&\mathcal{F}$,
    xlabel = FPS,
    ylabel shift = 1 pt,
    minor tick num=5,
    grid=both,
    grid style={line width=.1pt, draw=gray!10},
    major grid style={line width=.2pt,draw=gray!50},]

\addplot[
    only marks,
    mark=*,mark options={color=blue},
    mark size=1.2pt,
    visualization depends on=\thisrow{alignment} \as \alignment,
    nodes near coords, 
    point meta=explicit symbolic, 
    every node near coord/.style={anchor=\alignment, color=black} 
    ] table [
     meta index=2 
     ] {
FPS  acc  name alignment
0.22 60.3 OSVOS\cite{osvos}  180
10.6 81.8 STM\cite{STM} 40
7.7 66.7 RGMP\cite{rgmp} -40
14.4 76.7 FRTM-VOS\cite{FRTM} 40
11.2 84.5 MiVOS\cite{MiVOS}  -40
20.2 85.3 STCN\cite{STCN} -40
41 72.3 Track-Seg\cite{Track-Seg} 220
5.6 82.9 CFBI+\cite{yang2020CFBIP} 0
37 72.3 TVOS\cite{zhang2020transductive} 40
34.5 58.4 CVOS\cite{tan2020real} -40
};

\addplot[
    only marks,
    mark=*,mark options={color=purple},
    mark size=1.2pt,
    visualization depends on=\thisrow{alignment} \as \alignment,
    nodes near coords, 
    point meta=explicit symbolic, 
    every node near coord/.style={anchor=\alignment, font=\bfseries, color=black} 
    ] table [
     meta index=2 
     ] {
FPS  acc  name alignment
33.7 82.4 STCN+CoVOS  190
25.5 82.2 MiVOS+CoVOS  173
23.8 80.5 STM+CoVOS 150
};

\node[anchor=west] (source) at (axis cs:11.2,84.5){};
\node (destination) at (axis cs:25.5,82.2){};
\draw[->](source)--(destination);

\node[anchor=west] (source) at (axis cs:20.2,85.3){};
\node (destination) at (axis cs:33.7,82.4){};
\draw[->](source)--(destination);

\node[anchor=west] (source) at (axis cs:10.6, 81.8){};
\node (destination) at (axis cs:23.8, 80.5){};
\draw[->](source)--(destination);

\end{axis}
\end{tikzpicture}
\caption{Comparison of VOS methods on the DAVIS 17 dataset. We double the speed of STM, MiVOS, and STCN with minor drops in accuracy. The other compressed video method
CVOS~\cite{tan2020real} achieves comparable speed but has a significant drop in accuracy. 
}
\label{fig:speed_acc}
\end{figure}

We aim to develop an accurate yet efficient \emph{VOS acceleration framework}. As our interest is in acceleration, it is natural to follow a propagation-based approach in which an (heavy) off-the-shelf base network is applied to only keyframes.  Acceleration is then achieved by propagating the keyframe segmentations and features to non-keyframes.  In our framework, we leverage the information from the compressed video bitstream, specifically, the motion vectors and residuals, which are ideal for an efficient yet accurate propagation scheme.  

Motion vectors are cheap to obtain -- they simply need to be read out from the bitstream.  However, they are also more challenging to work with than optical flow.  Whereas optical flow fields are dense and defined on a pixel-wise basis, motion vectors are sparse. For example in HEVC~\cite{HEVC}, they are defined only for blocks of pixels, 
which greatly reduces the resolution of the motion information and introduces block artifacts. Furthermore, in cases where the coding bitrate limit is too low, the encoder may not estimate the motion correctly; this often happens in complex scenes or under fast motions. As such, we propose a dedicated soft propagation module that suppresses noise.  
For further improvement, we also propose a mask correction module based on the bitstream residuals. Putting all of this together, we designed a new plug-and-play framework based on compressed videos to accelerate standard VOS  methods~\cite{STM,MiVOS,STCN}.  We use these off-the-shelf methods as base networks to segment keyframes and then leverage the compressed videos' motion vectors for propagation and residuals for correction. 

A key distinction between our motion vector propagation module and existing optical flow propagation methods~\cite{dff,paul2020efficient,masktrack,rgmp} is that our module is bi-directional. We take advantage of the inherent bi-directional nature of motion vectors and propagate information both forwards and backwards. Our module is also multi-hop as we can propagate mask between non-keyframes. These features make our propagation scheme less prone to drift and occlusion errors.  

A closely related work to ours is 
CVOS~\cite{tan2020real}.  CVOS aims to develop a stand-alone VOS framework based on compressed videos, whereas we are proposing a plug-and-play acceleration module. A shortcoming of CVOS is that it considers only I- and P-frames but not B-frames in their framework. This setting is highly restrictive and uncommon, since B-frames were introduced to the default encoding setting specified by the MPEG-1 standard \cite{le1991mpeg} over 30 years ago. In contrast, we consider I-, P- and B-frames, making our method more applicable and practical for modern compressed video settings.

Our experiments demonstrate that our module offers considerable speed-ups on several image sequence-based models (see \cref{fig:speed_acc}). As a by-product of the keyframe selection, our module also reduces the memory of existing memory-networks~\cite{STM,KMN}, which are some of the fastest and most accurate state-of-the-art VOS methods. We summarize our contributions below:

\begin{itemize}
    \item A novel VOS acceleration module that leverages information from the compressed video bitstream for segmentation mask propagation and correction.
    \item A soft propagation module that takes as input inaccurate and blocky motion vectors but yields highly accurate warps in a multi-hop and bi-directional manner.
    \item A mask correction module that refines  propagation errors and artifacts based on motion residuals.
    \item Our plug-and-play module is flexible and can be applied to off-the-shelf VOS methods to achieve up to 3.5$\times\!$ speed-ups with negligible drops in accuracy.  
\end{itemize}

\section{Related work}\label{sec:related_work}
\noindent \textbf{Video object segmentation} approaches are either semi-supervised, in which an initial mask is provided for the video, or unsupervised, in which no mask is available.  We limit our discussion here to semi-supervised methods. 
Semi-supervised VOS methods can be further divided into two types: matching-based and propagation-based.
Matching-based VOS methods rely on limited appearance changes to either match the template and target frame or to learn an object detector. For example,~\cite{osvos,OnVOS,FRTM} ﬁne-tune a segmentation network using provided and estimated masks with extensive data augmentation.  Other examples include memory-networks~\cite{STM,KMN,MiVOS,STCN} that perform reference-query matching for the target object based on features extracted from previous frames. Propagation-based VOS methods rely on temporal correlations to propagate segmentation masks from the annotated frames.  A simple propagation strategy is to copy the previous mask~\cite{masktrack}, assuming limited change from frame-to-frame.  Others works use motion-based cues from optical flow~\cite{tsai2012motion, cheng2017segflow,hu2018motion}.

\noindent \textbf{Keyframe propagation.}
Frame-wise propagation of information from keyframes to non-keyframes has been used for efficient semantic video segmentation~\cite{dff,accel,paul2020efficient}, but little has been explored for its role in efficient VOS~\cite{key_segmentation} due to several reasons. Firstly, selecting keyframes is non-trivial. For maximum efficiency, keyframes should be as few and distinct as possible; yet if they are too distinct, the gap becomes too large to propagate across. As a result, existing works select keyframes conservatively with either uniform sampling~\cite{dff,accel} or thresholding of changes in low-level features~\cite{low_latency}.  Secondly, frame-wise propagation relies on optical flow, and computing accurate flow fields\cite{flownet2,raft} is still computationally expensive.

Our proposed framework is propagation-based, but we differ from similar approaches in that we use the compressed video bitstream for propagation and correction. Our method adaptively selects key-frames, and it is also the first to use a bi-directional and multi-hop propagation scheme. 

\noindent \textbf{Compressed videos} have been used in various vision tasks. Early methods~\cite{babu2004video,porikli2009compressed} used the compressed bitstream to form feature descriptors for unsupervised object segmentation  and detection.  In contrast, we utilize the bitstream for propagation and correction to accelerate semi-supervised VOS.  More recently, the use of compressed videos has been explored for object detection~\cite{wang2019fast}, saliency detection~\cite{hevc_saliency}, action recognition~\cite{CoViAR} and, as discussed earlier, VOS~\cite{tan2020real}. These works leverage motion vectors and residuals as motion cues or bit allocation as indicators of saliency. As features in the bitstream are inherently coarse, most of the previous works have a significant accuracy drop compared to methods that use full videos or optical flow.  Our work is the first compressed video method that can fill this gap.

\section{Preliminaries}
\subsection{Compressed video format}\label{compressedvideo}
Video in its raw form is a sequence of RGB images; however, it is unnecessary to store all the image frames. Video compression encoder-decoders, or codecs, leverage frame-to-frame redundancies to minimize storage. We outline some essentials of the HEVC codec~\cite{HEVC}; other codecs like MPEG-4~\cite{sikora1997mpeg} and H.264~\cite{AVC} follow similar principles.  Note that this section introduces only concepts relevant to understanding our framework. We refer to~\cite{hevcbook} for a more comprehensive discussion.

\begin{figure}[t!]

    \begin{subfigure}[b]{\linewidth}
        \centering
        \includegraphics[width=\textwidth]{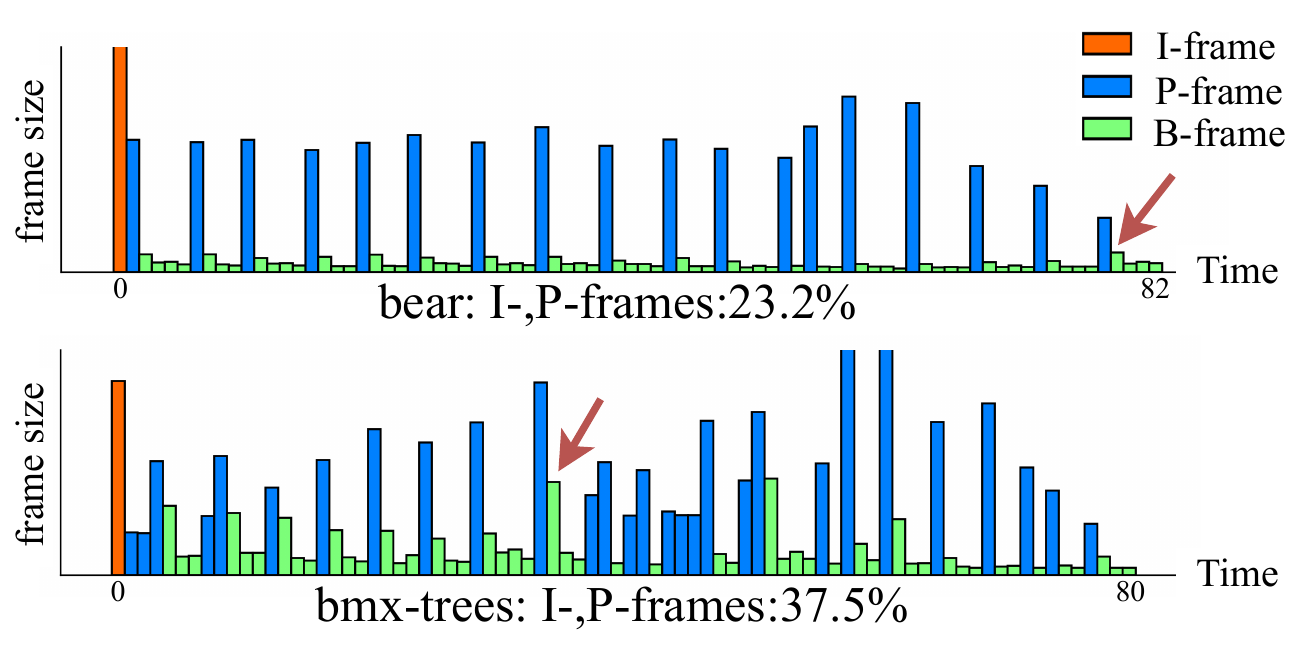}
    \end{subfigure}
    ~
        \begin{subfigure}[b]{\linewidth}
        \centering
        \includegraphics[width=0.8\textwidth]{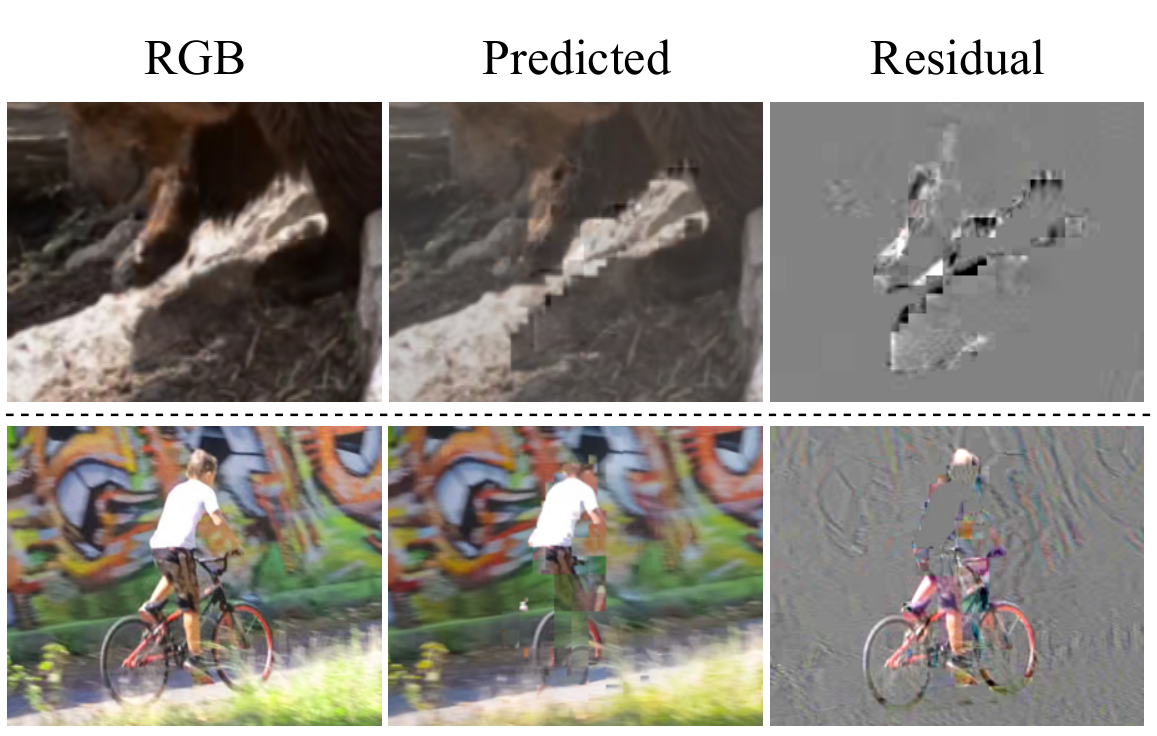}
    \end{subfigure}
    \caption{ Bar plots of a GOP visualizing frame assignments and the relative frame size. The \emph{`bmx-trees'} sequence has faster movements so it has more I-/P-frames than \emph{`bear'} (37.5\% vs. 23.2\%). The red arrows mark displayed frames, which feature examples of block effects for the \emph{`bear'} sequence above and motion vector estimation failures for the \emph{`bmx-trees'} sequence below.}\label{fig:frame_type}\label{fig:predicted_blockeffect}

\end{figure}

The HEVC coding structure consists of a series of frames called a Group of Pictures (GOP). Each GOP uses three frame types: I-frames, P-frames and B-frames. I-frames are fully encoded standalone, while P- and B-frames are encoded relatively based on motion compensation from other frames and residuals. Specifically, the P- and B-frames store motion vectors, which can be considered a block-wise analogue of optical flow between that frame and its' reference frame(s). Any discrepancies are then stored in that frame's residual. \cref{fig:frame_type} shows the frame assignments of two sample GOPs. Video decoding is therefore an ordered process to ensure that reference frames are decoded first to preserve the chain of dependencies. \cref{fig:mvdemo} illustrates the dependencies in a sample sequence.

\subsection{Motion compensation in compressed videos} \label{motionprediction}
A key difference between optical flow and motion vectors is that optical flow is a dense vector field with respect to a neighbouring frame in time, whereas motion vectors are block-wise displacements with respect to arbitrary reference frame(s) within the GOP. The associated blocks are called Prediction Units (PU), and they vary in size from $64\!\times\!64$ to $8\!\times\!4$ or $\!4\times\!8$ pixels. PUs can be uni-directional, with reference frames from either the past or the future, or bi-directional, 
with references to both the past and the future. 
P-frames have only uni-directional PUs, while B-frames have both uni-directional and bi-directional PUs. 

In this work, we denote a PU as $\Omega_{ij}$, with constituent pixels $(x,y)\in\Omega_{ij}$\footnote{For simplicity, we abuse notation and refer to both the PU and the constituent pixels simply as $\Omega_{ij}$.}, where $i$ indexes the frame and $j$ indexes the PU in frame $i$. In the general bi-directional case, $\Omega_{ij}$ is associated with a pair of forward and backward motion vectors $(\rbv_{ij}, \lbv_{ij})$, where the right and left arrows denote forward and backward motion, respectively.  The forward motion vector $\rbv_{ij} = [\ru, \rv, \rt\,]$ is given by displacements $\ru$ and $\rv$ and reference frame $\rt$, where $\rt\!<\!i$; analogously, $\lbv_{ij} = [\lu, \lv, \lt\,]$ denotes a backward motion vector with displacements $[\lu, \lv]$ and reference frame $\lt$, where $\lt\!>\!i$.

Based on the motion vectors, the pixels $(x,y)\!\in\!\Omega_{ij}$ can be predicted from co-located blocks of the same size as $\Omega_{ij}$ from reference frames $I_{\rt}$ and $I_{\lt}$. The reconstructed frame $\hat{I}_{i}^{x,y}$ at $(x,y)$ of frame $i$, for $(x,y)\in\Omega_{ij}$, is given as 
\begin{equation}\label{eq:bidir_recon}
\hat{I}_i^{x,y} = 
\rw I_{\rt\,}^{x+\ru,y+\rv}
+ \lw I_{\lt}^{x+\lu,y+\lv},
\end{equation}
\noindent where $(\rw,\lw)$ are weighting components for the forward and backward motions, respectively, and $\rw\!+\!\lw\!=\!1$.  In the case of a uni-directional PU, either $\rw$ or $\lw$ would be set to $0$ and the corresponding $\ovr{\bv}$ or $\ovl{\bv}$ is undefined. 

\begin{figure}[t!]
    \centering
    \includegraphics[width=\linewidth]{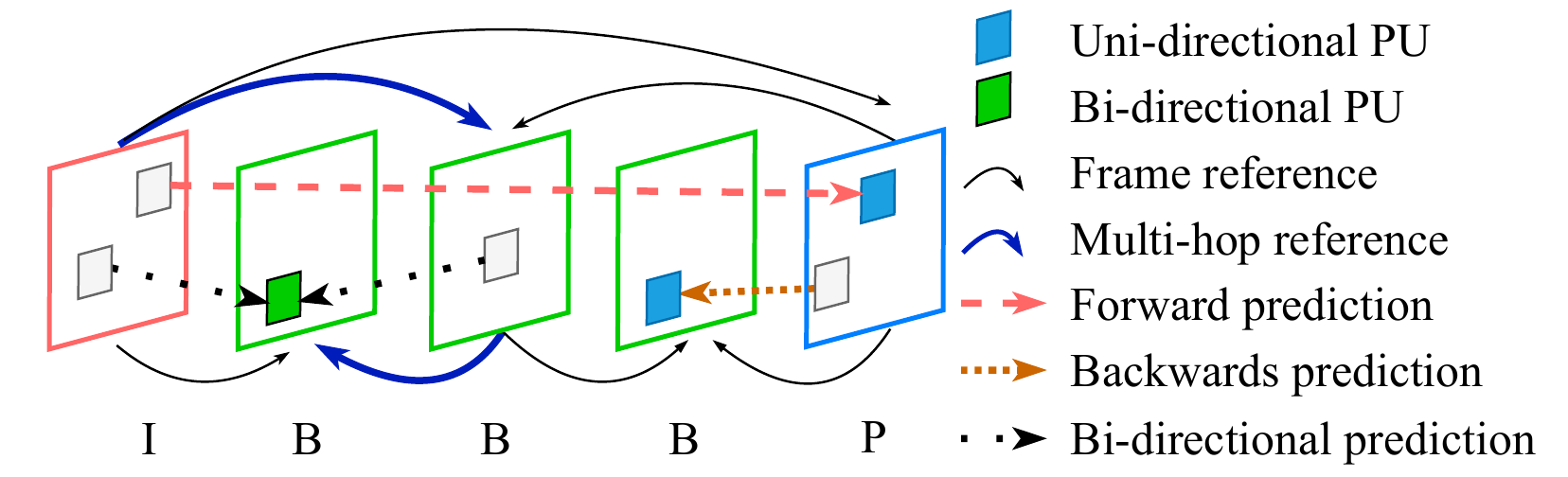}
    \caption{GOP schematic. Dashed lines denote motion compensation in prediction blocks. \emph{'I'}, \emph{'B'} and \emph{'P'} denote frame types. 
    }\label{fig:mvdemo}
\end{figure}

In older and more restrictive codec settings, such as those used in CVOS\cite{tan2020real}, reference frames were limited to I-frames. Modern codecs like HEVC, \ie what we consider in this work, allow P- and B-frames to reference pixels in other P- and B-frames, which are themselves reconstructed from other references. This makes the reconstruction in~\cref{eq:bidir_recon} \emph{multi-hop}, which improves overall coding efficiency as the drifting problem can be alleviated with smaller temporal reference distance. 
Examples of PUs and frame predictions are illustrated in~\cref{fig:mvdemo}.
Motion vectors are inherently coarse and noisy, due to their block-wise nature and encoding errors in areas of fast and abrupt movements (see examples in~\cref{fig:predicted_blockeffect}).
As such, the remaining differences between the RGB image $I_i$ and prediction $\hat{I}_i$ at frame $i$ are stored in the residual $\textbf{e}_i$ to recover pixel-level detailing:
\begin{equation}\label{eq:residual}
    I_i = \hat{I}_i + \textbf{e}_i.
\end{equation}
In principle, $\textbf{e}_i$ is sparse; the sparsity is directly correlated with the accuracy of the motion vector prediction. The key to efficient video encoding is balancing the storage savings of using larger PUs for P- and B-frames, \ie~fewer motion vectors, versus requiring less sparse residuals to compensate for the coarser block motions.

\subsection{Dense frame-wise motion representation}
Performing frame-wise propagation directly from the motion vectors can be cumbersome as the vectors are defined block-wise according to PUs. The PUs in a given frame often have several (different) references over multiple hops. As such, we compute a dense frame-wise motion field to serve as a more convenient intermediate representation. Specifically, we define a bi-directional motion field as $M_i\!=\![\rM_i, \lM_i]$, where $\rM_i\!\in \!\mathbb{R}^{H\times W\times 3}$ is a dense pixel-wise representation of forward motions for frame $i$ and is represented by $[\ru,\rv,\rt\,]$, \ie the displacements and the reference frame. Similar to the motion vectors, the right- and left-arrowed accents denote forward and backward motions respectively.  As such, $\lM_i\!\in\!\mathbb{R}^{H\times W \times 3 }$ stores backward motions for frame $i$ represented by $[\lu,\lv,\lt\,]$. The motion components are determined by aggregating all the PUs $\{\Omega_{ij}\}, j\in\{1...J_i\}$, where $J_i$ is the total number of PUs in frame $i$. \ie
\begin{multline}\label{eq:mv_field}
    \rbv_{\Omega, i} \rightarrow \rM_i^{x,y}; \quad
    \lbv_{\Omega, i} \rightarrow \lM_i^{x,y},\quad (x,y) \in \Omega_{ij} .
\end{multline}
This assignment procedure, which is denoted by $\rightarrow$, iterates through all the spatial locations of frame $i$. If a given PU in the B-frame is uni-directional, then the elements in the opposite direction in either $\ovr{M}$ or $\ovl{M}$ is set to zero accordingly. For pixels where $\lt$ or $\rt$  is directed to a keyframe, the prediction is single-hop; for pixels where $\lt$ or $\rt$ is directed to another non-keyframe, this will be multi-hop as the current reference is chained to further references.

\begin{figure*}[t!]
    \centering
    \includegraphics[width=0.9\textwidth]{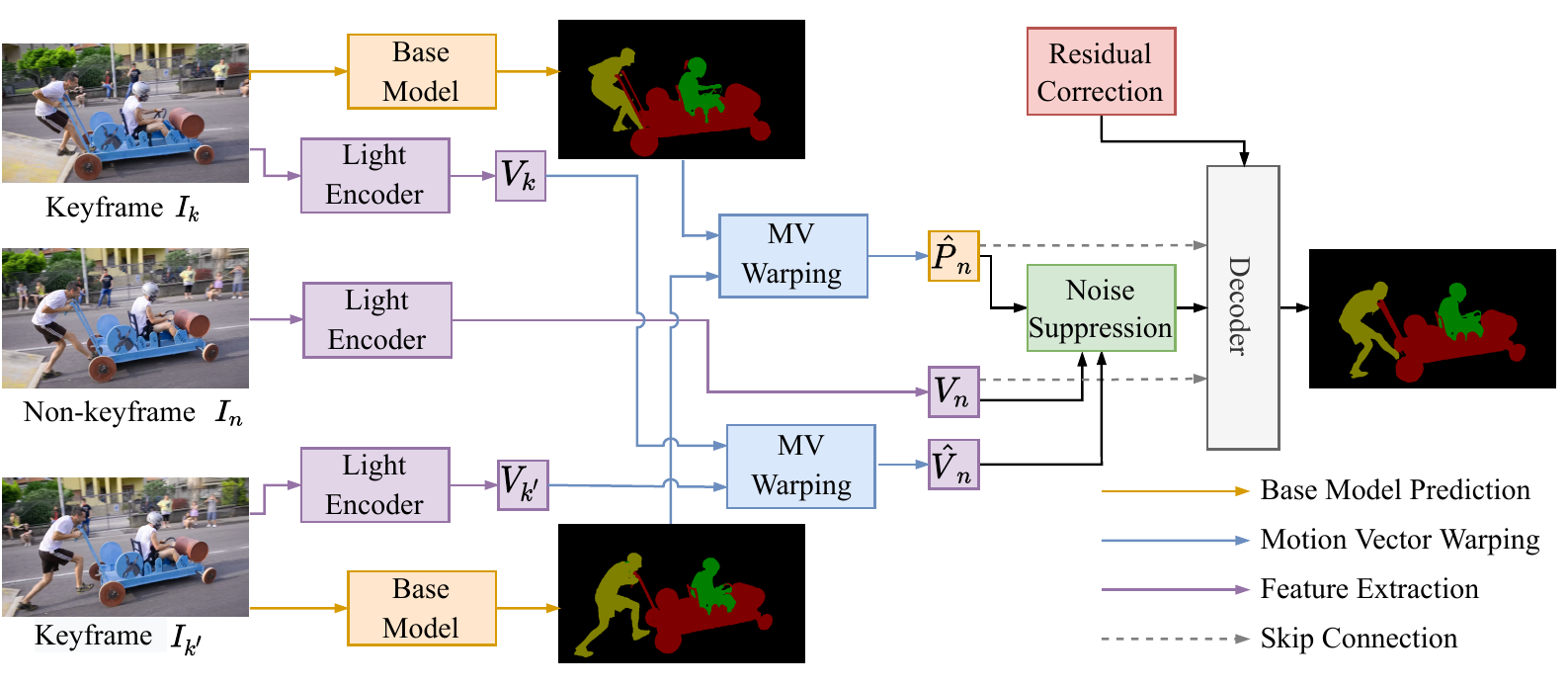}
    \caption{\textbf{Overall framework}. Keyframe segmentation predictions are propagated to non-keyframes through a soft motion vector propagation module that suppresses inaccurate motion vectors. Propagated masks are then corrected based on the residuals and feature matching.}\label{fig:framework}
\end{figure*}

\section{Methodology}
We accelerate off-the-shelf VOS methods by applying these methods as a base network to selected keyframes (\cref{sec:basekeyframe}).  The keyframe segmentations are propagated to non-keyframes with a soft motion vector propagation module (\cref{sec:propagation}) and further refined via a residual-based correction module (\cref{sec:correction}). ~\cref{fig:framework} illustrates the overall framework. The acceleration comes from the computational savings of propagation and correction compared to applying the base network to all frames in the sequence.

\subsection{Problem formulation}
We denote the decoded sequence from a compressed video bitstream of length $T$ as $\{(I_i,M_i,\textbf{e}_i), i \in [1, T]\}$. 

For convenience, we directly use the motion field $M_i$ instead of the raw motion vectors. Note that after decoding, we already have access to the RGB image $I_i$ for frame $i$.  For $P$ and $B$ frames, $I_i$ is reconstructed from the motion-predicted frame $\hat{I}_i$ and the residual $\textbf{e}_i$ based on~\cref{eq:residual}. 
For clarity, we maintain two redundant frame indices $n$ and $k$ for referring to non-keyframes and keyframes, respectively.
We denote the base network as $\{F, G\}$. The first portion of the network $F$ extracts low-level appearance features $V_k$ from the input keyframe $I_k$; $G$ denotes the subsequent part of the network that further processes $V_k$ to estimate the segmentation $P_k$, \ie, for a keyframe $k$, 
\begin{equation}\label{eq:base_net}
V_k = F(I_k), \;\; P_k = G(V_k),
\end{equation}
\noindent where $P_k \in \mathbb{R}^{ H \times W \times O} $ and $V_k \in \mathbb{R}^{H \times W \times C} $.  Here, $O$ is the number of objects in the video sequence, $C$ is the number of channels for the low-level feature and $H\times W$ is the spatial resolution of the prediction. 

For a non-keyframe $I_n$, a standard approach~\cite{dff} to
propagate the segmentation predictions from a keyframe $k$ is to apply a warp based on the optical flow: 
\begin{equation}\label{eq:basicwarp}
    \tilde{P}_n = W(\text{OF}_n,P_k), 
\end{equation}
\noindent where $W$ is the warping operation, $OF_n$ is the optical flow between $P_n$ and $P_k$, and $\tilde{P}$ is the propagated predictions. This form of propagation has two key drawbacks. Firstly, most schemes use optical flow computed only between two frames, which increases the possible errors that arise from occlusion. Secondly, estimating accurate optical flows still comes with considerable computational expense. 

\subsection{Soft motion-vector propagation module }\label{sec:propagation}
In this section, we outline how motion vectors, specifically the motion vector field $M_n$ defined in \cref{eq:mv_field} for a non-keyframe $I_n$, can be used in place of optical flow $\text{OF}_n$ in \cref{eq:basicwarp}. We first introduce the motion vector warping operation, in which $\hat{P}_n$ and $\hat{V}_n$ denote the motion vector warped prediction and warped features, \ie
\begin{equation}\label{eq:warp_prediction}
\hat{P}_n = W_{MV}(M_n,P_{\star}),\quad
\hat{V}_n = W_{MV}(M_n,V_{\star}),
\end{equation}

\noindent where $P_{\star}$ and $V_{\star}$ denote the corresponding segmentations and features for key- and non-key reference frames, respectively. The warping operation $W_{MV}$ is defined as a backward warp which iterates over all the spatial locations of frame $n$. If we denote with $\Lambda$ the item, \ie $P_{\star}$ or $V_{\star}$, to be propagated, such that $\hat{\Lambda}_n=W_{MV}(M_n,\Lambda)$, then the propagated value at $(x,y)$ for a non-keyframe $n$, based on~\cref{eq:bidir_recon} can be defined as:
\begin{equation}\label{eq:mv_warp}
  \hat{\Lambda}_n^{x,y} = 
\begin{cases}
\Lambda_{\lt}^{x+\lu, y+\lv}
  , \quad & \text{if}\; \rt = 0, \\
\Lambda_{\rt}^{x+\ru, y+\rv}
  , \quad &  \text{if}\; \lt = 0, \\
  \frac{1}{2}\Lambda_{\lt}^{x+\lu, y+\lv} + \frac{1}{2} \Lambda_{\rt}^{x+\ru, y+\rv}, \quad &  \text{otherwise}.
\end{cases}  
\end{equation}
\begin{equation}\label{eq:mv_element}
\text{where:} \qquad \qquad [\ru,\rv,\rt,\lu,\lv,\lt\;]=M_n^{x,y}.
\end{equation}

\noindent The first two cases in ~\cref{eq:mv_warp} are for warping unidirectional motion vectors forwards and backwards in time, respectively, and the third case is used for bi-directional motion vectors.  Note that in the third case, the forward and backward motion vectors are equally weighted and not according to $\rw$ and $\lw$ from \cref{eq:bidir_recon}.  This is because we interpret the references to be equally indicative of the target mask; also, $\rw$ and $\lw$ are tuned for reconstructing the target RGB pixel value. In the case when $u$, $v$ are not integers, nearest-neighbours or bilinear interpolation will be applied in the reference map; for simplicity, we omit the interpolation in the formulation.  If the reference frame $\lt$ or $\rt$ is not a keyframe, then the warping becomes multi-hop.  Hence, the warping procedure must follow the decoding order, as referenced non-keyframes must be completed before it can be propagated onwards.   
To mitigate the impact of noise and errors in the motion vector field, we propose a soft propagation scheme that makes use of a learned decoder $\mathcal{D}(\cdot)$: 
\begin{equation}\label{eq:soft_propagation}
   \tilde{P}_n = \mathcal{D}
   \left([\hat{P}_n,V_n, S(V_n,\hat{V}_n)\cdot\hat{P}_n] \right),
\end{equation}
\noindent where the square braces $[,]$ denote concatenation. The decoder is lightweight, and denoises the originally propagated mask $\hat{P}_n\!=\!W_{MV}(M_n,P)$ based on the low-level features of the input frame $I_n$, \ie $V_n\!=\!F(I_n)$, and a confidence-weighted version of the propagated mask. The weighting term $S(V_n,\hat{V}_n) \in \mathbb{R}^{H \times W}$ is defined by a similarity between the extracted features $V_n \in \mathbb{R}^{H \times W \times C}$ and the propagated features $\hat{V}_n \in \mathbb{R}^{H \times W \times C} $. We use dot product along the channel dimension to represent the similarity, \ie
\begin{equation}\label{eq:motion_vector}
S(V_n,\hat{V}_n)^{ij} = \sigma\,( V_n^{ij} \cdot \hat{V}_n^{ij}),
\end{equation}
where $\sigma$ is the standard sigmoid function. 
The similarity between the propagated features $\hat{V}_n$ and the actually estimated features $V_n$ serves as a confidence indicator to the decoder where the propagation is likely accurate. In areas which are not similar, the motion vector is likely inaccurate, so the propagated values should likely be suppressed and require more denoising.

\subsection{Residual-based correction module}\label{sec:correction}
We introduce an additional correction module to further improve the quality of the propagated segmentation masks. As errors of the motion vectors are captured inherently in each frame's residuals, it is natural to use these as a cue for compensation. We choose to model such correction through patching generation and label matching explicitly. While implicitly adding residual to the decoder network could achieve similar performance, it requires relatively more data and a heavier decoder network.

Let $\textbf{e} \in \mathbb{R}^{H \times W \times 3 }$ and $\hat{\mathbf{S}}$\footnote{$\textbf{e}_n$ and $\hat{\mathbf{S}}_n$ for completeness but we drop the subscript $n$ for simplicity}
denote the residuals and the propagated foreground mask, where $\hat{\mathbf{S}}$ can be obtained by taking \emph{argmax} of propagated prediction $\hat{P}$.
We first convert $\textbf{e}$ into a greyscale image before converting it into a binary mask $\textbf{e}_b$ via thresholding.  The corrected mask $\tilde{\textbf{S}}$ is found by taking the intersection between $\textbf{e}_b$ and $\hat{\mathbf{S}}_{+}$, a dilated version of initially propagated mask $\hat{\mathbf{S}}$, \ie,  
$\tilde{\mathbf{S}} = \cap( \textbf{e}_b,\hat{\mathbf{S}}_{+})$,
where $\cap(\cdot)$ indicates an intersection operation and allows us to focus only on foreground areas of the dilated mask, which coincide with thresholded residual values. 

$\tilde{\mathbf{S}}$ provides an indication of which areas in the propagated mask will require correction. For each pixel in $\tilde{\mathbf{S}}$ indexed by $a$ at frame $n$, we search in the temporally closest keyframe $k^*$ and match between $V_{n}$ and $V_{k^*}$. Specifically, we define $\mathbf{W}^{ak}$ as the affinity between the feature at pixel $a$ in $V_{n}$, \ie $V^a_{n}$, and all pixels in $V_{k^*}$. The corrected mask prediction at pixel $a$ is then obtained by $P^a_n = \mathbf{W}^{ak} P_{k^*}$. We use an L2-similarity function to compute the affinity matrix and defer the details to the Supplementary.

\begin{figure}[t!]
    \centering
    \includegraphics[width=\linewidth]{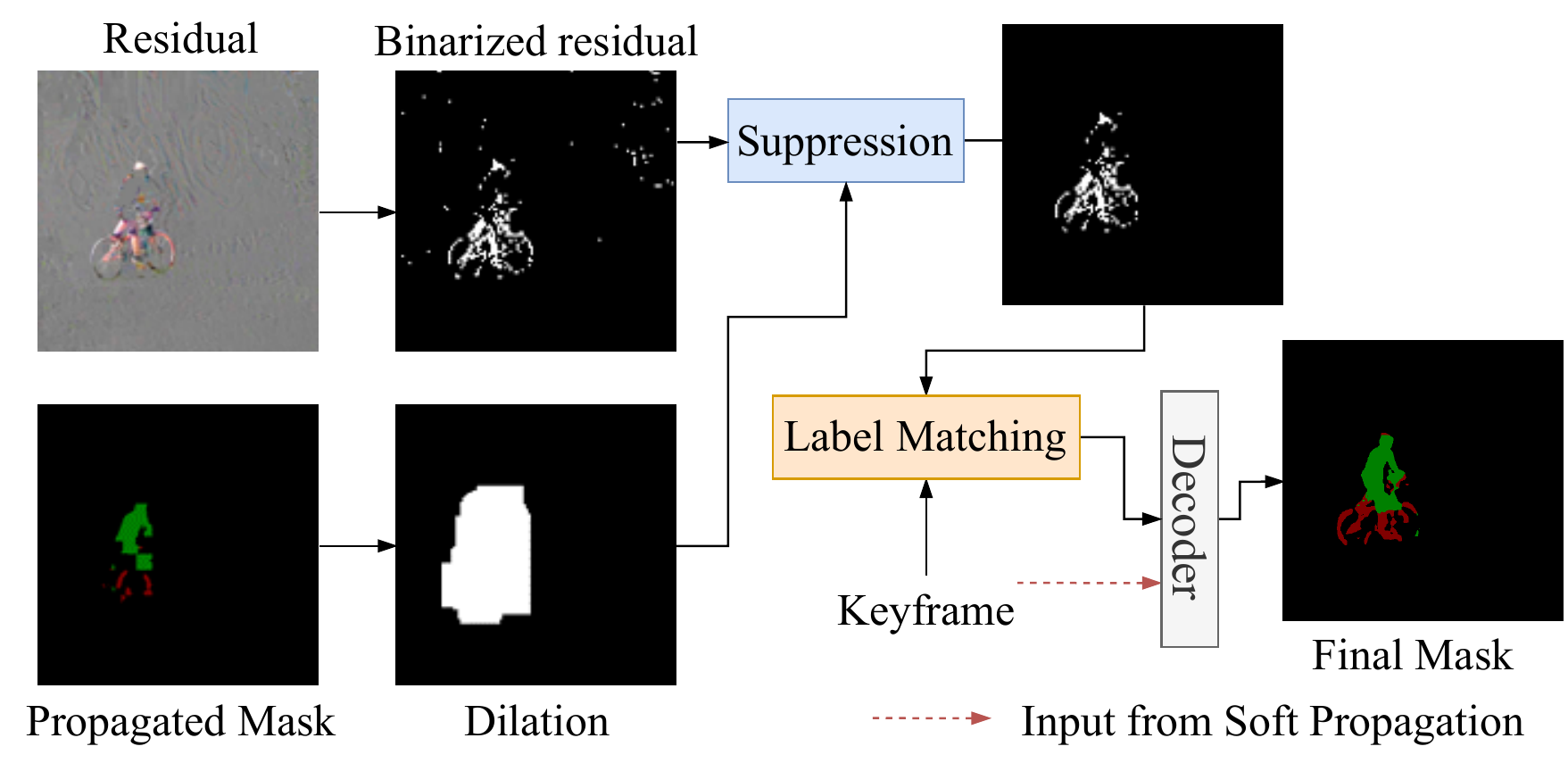}
    \caption{Residual-based correction module selects pixels to correct in the propagated mask; the correction scheme replaces the segmentation labels via a feature matching scheme.}
    \label{fig:residual}
    \vspace{-0.5em}
\end{figure}

\subsection{Keyframe \& base network selection}\label{sec:basekeyframe}

In principle, any frame can be a keyframe. However, it is natural to define keyframes according to the compressed frame type, as the encoder designates types based on the video's dynamic content. 
In addition to I-frames, we also choose P-frames as keyframes. This is because less than 5\% of frames in a video sequence are I-frames in the default HEVC encoding, which is insufficient for accurate propagation,  so we also include the 15-35\% of frames designated as P-frames. Considering P-frames as keyframes also helps improve the accuracy because the motion compensation in P-frames is strictly uni-directional.  Otherwise, propagation to these frames may suffer inaccuracies arising from occlusions in the same manner as optical flow.

For a base VOS model to be accelerated, most matching based segmentation models discussed in ~\cref{sec:related_work} are suitable as they rely only on the appearance of the target object. From preliminary experiments, we observed that VOS methods that use memory-networks such as STM~\cite{STM}, MiVOS~\cite{MiVOS}, and STCN~\cite{STCN} are ideal for acceleration. This is because the choice of using I- and P-frames as keyframes naturally aligns with the memory concept and allows for the selection of a (even more) compact yet diverse memory.  
\begin{table*}[t!]
  \footnotesize
  \centering 
  \caption{Comparison of acceleration on different base models with state-of-the-art methods. \dag Frame rates were measured on our device if originally not provided; we also re-estimated STM time on our hardware as we obtained higher FPS than their reported value. FPS on Youtube-VOS  is measured on the first 30 videos.}\label{tab:sota}
    \begin{tabular}{lcccc|cccc|cccccc}
    \hline
    &\multicolumn{4}{c|}{DAVIS16 validation}&\multicolumn{4}{c|}{DAVIS17 validation}&\multicolumn{6}{c}{YouTube-VOS 2018 validation} \\ 
    
    Method & $\J$ & $\F$ & $\J\&\F$ & FPS& $\J$ & $\F$ & $\J\&\F$& FPS & $\mathcal{G}$ & $\J_s$ & $\F_s$ & $\J_u$ & $\F_u$ &FPS  \\     
    \hline
    CVOS\cite{tan2020real} & 79.1& 80.3 &79.7 & 34.5 &57.4&59.3&58.4& 31.2 & -& -& - &- &- &-\\ 
    TVOS\cite{zhang2020transductive} & -& - &- & - &69.9&74.7&72.3& 37 & 67.8 &67.1 &69.4 &63.0 &71.6 &-\\
    Track-Seg\cite{Track-Seg} & 82.6& 83.6 &83.1 & 39 &68.6&76.0&72.3& $<$39 & 63.6& 67.1& 70.2 &55.3 &61.7 &-\\ 
    PReMVOS \cite{premvos}  & 84.9 & 88.6 & 86.8 & 0.03 &  73.9 &  81.7   &  77.8 &$<$0.03  & 66.9 &71.4 &75.9 &56.5& 63.7& -  \\ 
    SwiftNet \cite{wang2021swiftnet}  &  90.5  & 90.3 & 90.4 & 25 &  78.3    &  83.9   &  81.1   & 25 &77.8 &77.8  & 81.8&72.3 &79.5 & -  \\ 
    CFBI+ \cite{yang2020CFBIP}  &  88.7 & 91.1 & 89.9 & 5.6 &  80.1 &  85.7   &  82.9   & $<$5.6 &82.0 &81.2 &86.0& 76.2 &84.6 & -  \\

    \hline
    FRTM-VOS \cite{FRTM} &  - & -    & 83.5 &  21.9    &  -   &  -   & 76.7 & \dag14.1 &   72.1 &72.3 &76.2 &65.9&74.1 & \dag7.7  \\
    FRTM-VOS  + CoVOS  & 82.3  & 82.2& 82.3 &  28.6  & 69.7  & 
    75.2  & 72.5& 20.6 &65.6 & 68.0&71.0 & 58.2 & 65.4 & 25.3 \\
    \hline
    STM \cite{STM}     & 88.7 &89.9 &  89.3 &\dag14.9 & 79.2 & 84.3 &  81.8 &  \dag10.6&  79.4 &79.7& 84.2 &72.8& 80.9 & -  \\
    STM  + CoVOS  & 87.0  & 87.3 & 87.2  & 31.5   & 78.3   & 82.7  & 80.5 & 23.8 & - & - & - & - & - & -  \\
    \hline
    MiVOS \cite{MiVOS}  &  89.7 & 92.4 & 91.0 & 16.9 & 81.7 & 87.4 & 84.5 &11.2 & 82.6 & 81.1 & 85.6 & 77.7 & 86.2 & \dag13   \\  
    MiVOS  + CoVOS & 89.0  & 89.8 &89.4 &36.8 & 79.7 & 84.6 & 82.2& 25.5 &79.3 &78.9 &83.0 &73.5 &81.7 &    45.9\\
    \hline
    STCN \cite{STCN}  &   90.4& 93.0& 91.7 & 26.9 & 82.0 & 88.6 & 85.3 &20.2 &84.3 & 83.2 & 87.9 & 79.0 & 87.3 &  \dag16.8 \\  
    STCN  + CoVOS  & 88.5 &89.6 & 89.1  & 42.7 & 79.7 & 85.1 & 82.4 & 33.7 & 79.0& 79.4&83.6 &72.6 &80.4 &  57.9 \\
    \hline
    
  \end{tabular}
  
\end{table*}

\section{Experimentation}
\subsection{Experimental settings}\label{implementationdetails}

\noindent\textbf{Video Compression.}
We generated compressed video from images using the x265 library in FFmpeg on the \textit{default} preset. To write out the bitstream, we modified the decoder from openHEVC\cite{openhevc1,openhevc2} and shared the code publicly to encourage others to work with compressed video.

\noindent\textbf{Datasets \& Evaluation.} 
We experimented with three video object segmentation benchmarks: DAVIS16\cite{davis2016} and DAVIS17\cite{davis2017}, which are small datasets with 50 and 120 videos of single and multiple objects, respectively, and 
YouTube-VOS\cite{ytvos}, a large-scale dataset with 3945 videos of multiple objects. We used the images in their original resolution for encoding the videos. The default HEVC encoding produced an average of \{37\%, 36\%, 27\%\} of I/P-frames, and therefore keyframes per sequence 
for DAVIS16, DAVIS17 and YouTube-VOS, respectively.

We evaluated with the standard criteria from~\cite{davis2016}: Jaccard Index $\mathcal{J}$ (IoU of the output segmentation with ground-truth mask) for region similarity, and mean boundary $\mathcal{F}$-scores for contour accuracy.  
Additionally, we report the average over all seen and unseen classes for YouTube-VOS.

\noindent\textbf{Propagation \& Correction.}
In our propagation scheme, we applied reverse mapping for warping and nearest-neighbour interpolation kernels. The decoder in the soft propagation (\cref{sec:propagation}) is a lightweight network of three residual blocks (see Supplementary for details). The decoder is trained from scratch, with a uniform initialization and a learning rate of 1e-4 with a decay factor of 0.1 every 10k iterations for 40k iterations. For residual-based correction, the binary threshold was set to $0.15\!*\!255$ for the absolute value of gray-scaled residual. 

\noindent\textbf{Base Models.}
We show experiments accelerating four base models: STM~\cite{STM}, MiVOS~\cite{MiVOS}, STCN~\cite{MiVOS} and FRTM-VOS~\cite{FRTM}.  The first three use a memory bank; for a fair comparison, we allow only keyframes to be stored in the memory bank. We set the memory frequency to 2 on DAVIS and 5 on Youtube-VOS, as the latter has higher frame rates. In the experiments, both settings reduced the memory bank size. We refer to Supplementary for the memory analysis. FRTM-VOS fine-tunes a network based on the labelled frame and associated augmentations.  We fed only the keyframes into the network for segmentation and fine-tuning. In practice, this is equivalent to segmenting a temporally reduced video.

\subsection{Acceleration on different base models.}
\cref{tab:sota} compares our accelerated results on the four base models with other state-of-the-art models.  Our method achieves an excellent compromise between accuracy and speed. 
On DAVIS16 ($\approx\!37\%\!$ keyframes), we achieved $1.3\times$, $2.1\times$, $2.2\times$, $1.6\times$ speed-ups with a minor drop of $\mathcal{J}\&\mathcal{F}$ $1.2$, $2.1$, $1.6$, $2.6$ on FRTM-VOS, STM, MiVOS and STCN, respectively.
On DAVIS17 ($\approx36\%\!$ keyframes), we achieved  $1.5\times$, $2.2\times$, $2.3\times$, $1.7\times$ speed-ups with the drop on $\mathcal{J}\&\mathcal{F}$ $4.2$, $1.3$, $1.7$, $2.9$ for the same order of models.   

On YouTube-VOS ($\approx\!27\%\!$ keyframes), we achieved $3.3\times$, $3.5\times$, $3.4\times$ speed-ups with a $4.8$, $2.4$, $4.0$ drop of $\J_s \& \F_s$ for 
FRTM-VOS, MiVOS and STCN, respectively. We have larger drops on $\J_u\&\F_u$ for unseen data because our decoder is not pre-trained on larger datasets.
Note that the video lengths of YouTube-VOS are relatively long ($>$150frames), so the above methods  
require additional memory or additional online fine-tuning, which allows us to achieve higher speed-ups. 
Moreover, the lower keyframe percentage of YouTube-VOS also provides more speed-ups. We do not provide the result on STM
as no pre-trained weights are available. 

With an STCN  
base model, our performance on DAVIS17 is 1.3 to 10.1 higher than other efficient methods SwiftNet~\cite{wang2021swiftnet}, TVOS~\cite{zhang2020transductive} and Track-Seg~\cite{Track-Seg} with comparable frame rates, though our success should also be attributed to the high STCN base accuracy. Another compressed video method CVOS~\cite{tan2020real} achieves comparable speed but has a significant accuracy gap. 

\subsection{Ablation studies}
We verified each component of our framework. All ablations used MiVOS\cite{MiVOS} as the base model on default video encode preset unless otherwise indicated.

\noindent\textbf{Propagation.} We first compare with optical flow as a form of propagation, and consider a forward unidirectional flow warping as done in 
\cite{flownet,masktrack,dff}, using the flow from the state-of-the-art method RAFT~\cite{raft} (\emph{`Optical Flow'}). We also consider a bi-directional optical flow warping (\emph{`Bi-Optical Flow'}), which is used in~\cite{pan2019video}.  Additionally, we compare with two motion vector baselines from a work on compressed videos, CoViAR~\cite{CoViAR} (\emph{`MV to Flow'}), and another work on compressed VOS, CVOS~\cite{tan2020real} (\emph{`MV I to P'}). CoViAR converts motion vectors into a flow between two frames,~\ie for motion vector field $M_i$ at $(x,y)$ and frame $i$, $M_{of}(x,y)=[u,v]/[(t-i)\cdot fps]$ is the motion of the pixel $(x,y)$ in the unit time on plane $i$. CVOS has a further simplified motion vector usage and references all motions from one I-frame in the GOP. We compare our bi-directional, multi-hop motion vector warping with (\emph{`MV Soft Prop'}) and without (\emph{`MV Warp'}) the soft propagation, which performs further noise suppression. 

\begin{table}[t]
 \footnotesize
\centering
 \caption{Comparison of propagation methods, \emph{'B', 'M', 'Sup'} denotes bi-directional, multi-hop and noise suppression, respectively. $^{\dag}$No code given, we report the results from earlier work~\cite{tan2020real}} 
 \label{tab:warping}
  \begin{tabular}{lp{0.7em}p{0.7em}p{0.7em}cc|cc}
    \hline
    && & &\multicolumn{2}{c|}{DAVIS16}&\multicolumn{2}{c}{DAVIS17}\\
    Method & B & M& Sup& $\mathcal{J}$  & $\mathcal{F}$& $\mathcal{J}$ &  $\mathcal{F}$    \\
    \hline
    Optical Flow & & &                                  & 77.4          & 79.2      & 71.5      & 77.6  \\
    Bi-Optical Flow~\cite{pan2019video} &\xmark & &     & 85.0          & 87.4      & 75.9      & 81.7  \\
    MV I to P~\cite{tan2020real} & & &                  & 31.5$^{\dag}$ & -         &-          &-      \\
    MV to Flow~\cite{CoViAR}  & & &                     & 77.2          & 80.2      & 69.4      & 76.3  \\
    MV Warp        & \xmark&\xmark &                    & 85.7          & 89.2      & 77.2      & 84.4   \\
    MV Soft Prop  & \xmark&\xmark & \xmark              & 89.0          & 89.8      & 79.7      & 84.6  \\
    \hline
    No propagation~\cite{MiVOS}&  & &                   & 89.7          & 92.4      & 81.7      & 87.4  \\ 
    \hline
  \end{tabular}
  \vspace{-1.5em}
\end{table}

\cref{tab:warping} verifies the effectiveness of our proposed propagation. Bi-directional optical flow, originally used for video generation~\cite{pan2019video}, performs better than unidirectional optical flow because it is less affected by occlusion. CoViAR~\cite{CoViAR} is a compressed video action recognition system; their propagation is on par with optical flow. The simplified case in CVOS~\cite{tan2020real} fails to propagate meaningful segmentation masks and thus relies on heavy refinement.

\begin{figure}[h]
    \centering
    \includegraphics[width=\linewidth]{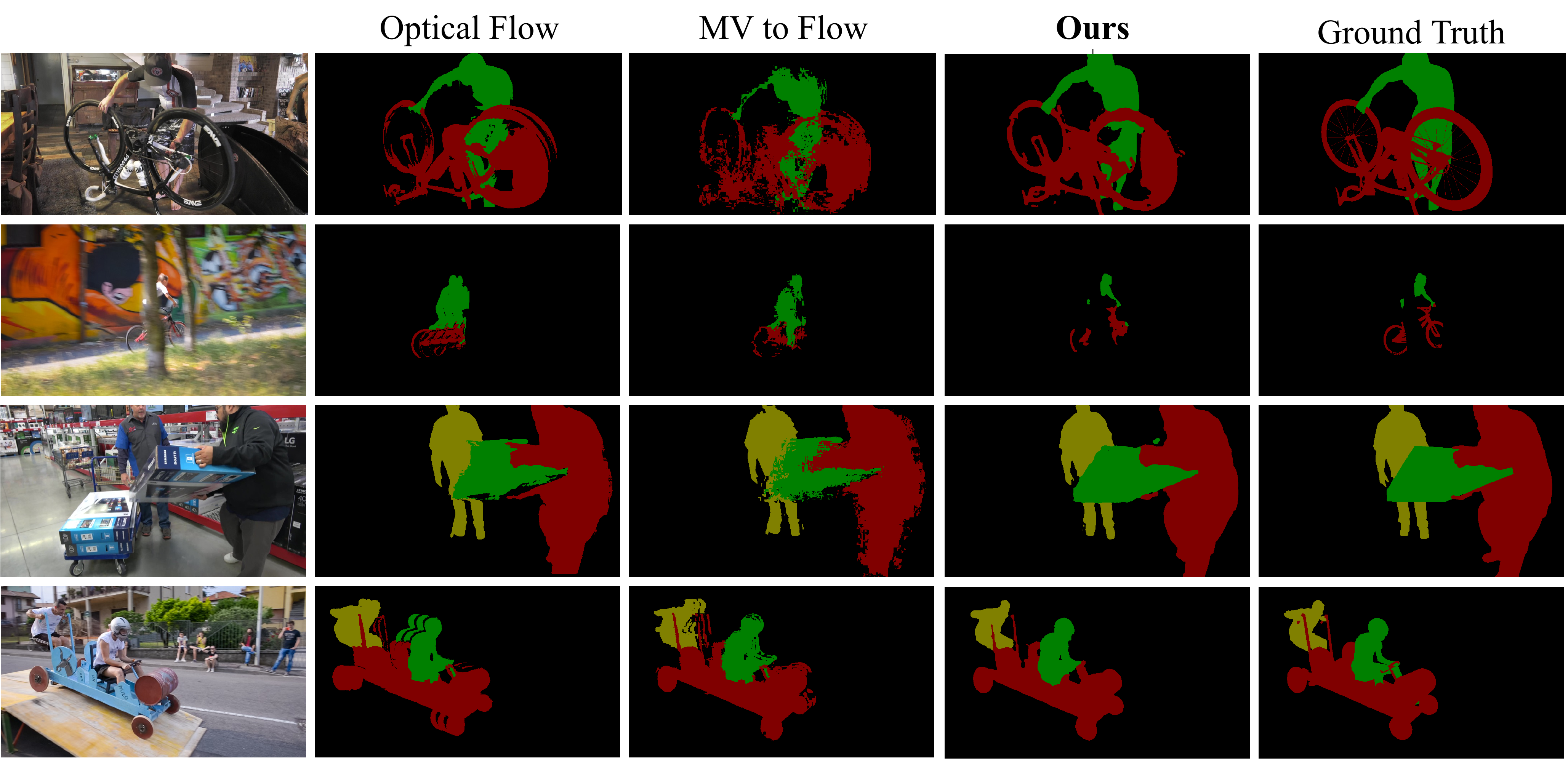}
    \caption{Optical flow propagation and motion-vector generated flows both suffer from ghosting effects and holes in areas of occlusion. Our propagation successfully prevents such artifacts. }
    \label{fig:qualwarp}
\end{figure}

Our bi-directional multi-hop motion vector-based warping outperforms all of the above methods.  Our soft propagation scheme with noise suppression gives further improvements to the accuracy, such that our propagated masks are within 4.0 points on both $\mathcal{J}$ and $\mathcal{F}$ below the upper bound without propagation, \ie by applying each frame through the base network. \cref{fig:qualwarp} shows qualitative results on different propagation methods.

\begin{table}[ht]
 \footnotesize
\centering
\caption{Ablations on decoder and mask correction module.}\label{tab:refinemodules}
\vspace{-0.5em}
  \begin{tabular}{lcc|cc}
    \hline
      &\multicolumn{2}{c|}{DAVIS16}&\multicolumn{2}{c}{DAVIS17}\\
    Module & $\mathcal{J}$  & $\mathcal{F}$& $\mathcal{J}$ &  $\mathcal{F}$    \\
    \hline
    MV Warp                 & 85.7 & 89.2 & 77.2 & 84.4\\
    +Decoder                & 88.3 & 88.8 & 79.2 & 84.0 \\
    +Suppression            & 88.8 & 89.6 & 79.6 & 84.5 \\
    +Residual Correction    & 89.0 & 89.8 & 79.7 & 84.6 \\
    \hline
  \end{tabular}
  
\end{table}

\noindent\textbf{Decoder and mask correction.} \cref{tab:refinemodules} shows how adding each component of the mask decoder leads to progressive improvements for the $\mathcal{J}$-index and boundary $\mathcal{F}$-score. For \emph{`MV Warp'}, we directly warp the prediction results on the original size of the frame. For the decoder, we warp the prediction and low-level features at 1/4 size for speed consideration. Because the motion vector is coarse and noisy, only input propagated prediction and the low-level features to the decoder will decrease accuracy.  The most significant gains come from the noise suppression module, \ie by feeding the suppressed propagated prediction into the decoder. Further residual correction increases the robustness for the corner cases.

\noindent\textbf{Keyframe percentage.}
To highlight the speed-accuracy trade-off, we compare the percentage of keyframes in \cref{tab:preset} by adjusting encoder presets.  The default HEVC setting yields $\approx37\%$ keyframes for DAVIS16 and DAVIS17. If we set the encoder to allocate more B-frames to have only approximately $25\%$ and $13\%$ keyframes (\emph{`B-frame biased’} and \emph{`Uniform B-frames'}, respectively), the propagated scores decrease while the FPS values increase accordingly. At the fastest setting, we can achieve 3.7x speed-ups on MiVOS with $\mathcal{J}\&\mathcal{F}$ scores of 82.9 on DAVIS16 and 4.5x speed-ups with $\mathcal{J}\&\mathcal{F}$ scores of 73.2 on DAVIS17.

\begin{table}[h]
 \footnotesize
\caption{Robustness to different video encoding presets on DAVIS16 and DAVIS17. B-frame biased: more weight on B-frame allocation (\emph{x265 option: bframe-bias=50}). Uniform B-frames: fixed 8 B-frames between I/P frames. }
\label{tab:preset}
\vspace{-0.5em}
\centering
\begin{tabular}{lc|cc|cc}
    \hline

    & &\multicolumn{2}{c|}{DAVIS16}&\multicolumn{2}{c}{DAVIS17}\\
    Preset& Keyframe & $\mathcal{J} \& \mathcal{F}$& FPS & $\mathcal{J}\&\mathcal{F}$& FPS \\
    \hline
    Default & $\approx37\%$ & 89.4 & 36.8 & 82.2&25.5\\

    B-frame biased 
    &$\approx25\%$ & 85.1&48.2 & 80.2&36.7 \\

    Uniform B-frames
    &$\approx13\%$ & 82.9&62.9 & 73.2&50.0 \\
    \hline
    No Propagation &-&91.0 &16.9 & 84.5&11.2\\
    \hline
  \end{tabular}
\end{table}

\subsection{Timing analysis}\label{timeanalysis}
To compute the FPS values in all our tables, we measured run times on an RTX-2080Ti for DAVIS dataset and on an RTX-A5000 for YouTube-VOS, as it requires extra memory. The amortized per frame inference time can be approximately computed by $T_{\text{base}}\!\cdot\!R\!+\!(T_{\text{propagation}}+\!T_{\text{correction}})\!\cdot\! (1\!-\!R)$, where $R$ denotes the ratio of keyframes.  Note that the measured $T_{\text{base}}$ may not correspond to the published FPS values of the base model, \eg for  
STM\cite{STM} and MiVOS\cite{MiVOS}. Our $T_{\text{base}}$ is lower because we store fewer frames in the memory bank (see Supplementary for more details). We measured the propagation and correction time on DAVIS17, and the sum ($T_{\text{propagation}}\!+\!T_{\text{correction}}$) is $12$ms.  

\section{Conclusion \& limitations}\label{conclusion}
We propose an acceleration framework for semi-supervised VOS via propagation by exploiting the motion vectors and residuals of the compressed video bitstream. Such a framework can speed up the accurate but slow base VOS models with minor drops in segmentation accuracy. One limitation of our work is the possible latency introduced by the multiple reference dependencies.  As a result, segmentation results of a non-keyframe get completed later than the future frame to which it refers. 

Given that 70\% of the internet traffic~\cite{index2017forecast} is dedicated to (compressed) videos, we see broad applicability of our work for acceleration.  
Efficiency in VOS methods is especially relevant for applications such as video editing,  given the growing trend of higher resolution videos,~\eg~ 4K standards. However,
VOS could also be abused to falsify parts of videos or create malicious content. We maintain a rigorous attitude towards this while emphasizing its positive impact on content creation and other possible improvements for the community.

\section{Acknowledgements}
This research is supported by the National Research Foundation, Singapore under its NRF Fellowship for AI (NRF-NRFFAI1-2019-0001). Any opinions, findings and conclusions or recommendations expressed in this material are those of the author(s) and do not reflect the views of National Research Foundation, Singapore.

{\small
\bibliographystyle{ieee_fullname}
\bibliography{reference}
}

\end{document}


\title{Accelerating Video Object Segmentation with Compressed Video}

\author{
 Kai Xu \quad Angela Yao\\
 National University of Singapore \\
 {\tt\small \{kxu, ayao\}@comp.nus.edu.sg}
}
\maketitle

\tableofcontents
\addcontentsline{toc}{section}{Algorithm Details}
\section*{Algorithm Details}
Given a compressed video sequence of length $T$, 
let $\{i, i\in[1,T]\}$ be the natural presentation index of the frames in time.  Additionally, we denote with $\mathbf{\Gamma} = \{\Gamma_i, i\in[1,T]\}$ the decoding order, where $\Gamma_i$ is the decoding index for frame $i$.  
Recall that $\{(I_i, M_i, \textbf{e}_i), i, i\in[1,T] \}$ is the reconstructed image sequence, where $I_i$ is the reconstructed RGB image of frame $i$ and  
$\{(M_i, \textbf{e}_i)\}$ are the associated motion fields and residuals.  $\mathcal{I}$, $\mathcal{P}$, $\mathcal{B}$ are sets of I-, P- and B-frame indices, respectively.
$W_{MV}(\cdot)$ is the motion vector warping operation defined in Eq.~6 and Eq.~7 of the main paper. 
The overall pipeline can be summarized in Algorithm 1.

\begin{algorithm}[t]
\small
\caption{Compressed video object segmentation.}
\begin{algorithmic}[1]
    \STATE \textbf{Model}: Base Model $\{F, G\}$, Decoder Model $\mathcal{D}$.
    \STATE \textbf{Input}: Decoding order $\{\Gamma_t, t\in[1,T]\}$; Reconstructed RGB $\{(I_i), i \in [1, T]\}$; Motion fields and residuals $\{(M_i,\textbf{e}_i), i \in [1, T]\}$; Residual binarization threshold $\tau$; $\mathcal{I}$, $\mathcal{P}$, $\mathcal{B}$ are sets of indices for I-, P- and B-frames respectively. $\psi$ denotes \emph{softmax aggregation} which is used in RGMP\cite{rgmp}. $\mathds{1}[\cdot]$ is indicator function.
\vspace{1em}
    \FOR{$i = \Gamma_1,...,\Gamma_T$}
        \IF{$i \in \mathcal{I}\ or\ \mathcal{P}$}
        \STATE Get key frame's softmax prediction from base model: $P_i = G(F(I_i))$, low level features: $V_i=F(I_i)$. 
       
        \ELSE   
        \STATE \textbf{Propagation:}
        \STATE \hspace{1em} Warp predictions: $\hat{P}_i = W_{MV}(M_i,P_*)$
        \STATE \hspace{2em} Buffer for multi-hop reference: $P_* \leftarrow \hat{P}_i$
        
        \STATE \hspace{1em} Warp features: $\hat{V}_i=W_{MV}(M_i, V_*)$
        \STATE \textbf{Confidence Re-weighting:}
        \STATE \hspace{1em} Compute features: $V_i=F(I_i)$
        \STATE \hspace{1em} Compute confidence-weighted prediction: 
        \\\hspace{2em} $\dot{P}_i=S(V_i, \hat{V}_i)\cdot \hat{P}_i $
        \STATE \textbf{Correction:}
        \STATE \hspace{1em}Binarize residual: $\textbf{e}_b = |greyscale(\textbf{e})| > \tau$
        \STATE \hspace{1em}Get warped foreground mask: \\ \hspace{2em}$\hat{\mathbf{S}}=\mathds{1}[argmax(\psi(\hat{P}))>0]$
        \STATE \hspace{1em}Get dilation of the foreground mask: \\
        \hspace{2em}$\hat{\mathbf{S}}_{+}=dilate(\hat{\mathbf{S}})$
        \STATE \hspace{1em}Get selected pixels to correct: 
        $\tilde{\mathbf{S}}=U(\textbf{e}_b, \hat{\mathbf{S}}_{+})$
        \STATE \hspace{1em} Feature matching: \\
        \hspace{2em} $\bar{P}_i=\text{feature-matching}(\tilde{\mathbf{S}}, P_{k*}, V_i, V_{k*})\!+\!\hat{P}_i$
        \STATE \textbf{Decoder:}
        \\\hspace{2em}$P_i=\mathcal{D}\big([\hat{P}_i, V_i, \dot{P}_i, \bar{P_i}] \big)$
        \ENDIF
        
    \ENDFOR
    \STATE 
    \STATE \textbf{Output}: All segmentations $\mathcal{S}=argmax(\psi(P_i))$

\end{algorithmic}
\end{algorithm}

\addcontentsline{toc}{section}{Additional Implementation Details}
\section*{Additional Implementation Details}

\textbf{Low-level feature encoder $F$:} With MiVOS~\cite{MiVOS} as the base model, we use the layers before the second residual stage of their RGB encoder (ResNet 50) as the low-level feature encoder. The encoder outputs features of 1/4 spatial size and 256 channels.

\textbf{Training:} All training data is generated with MiVOS as the base model. We generate and store the warped features, warped predictions, and residuals in advance using the weights for the model released by the authors. We use a pixel-wise \emph{cross entropy} loss for training. The training is performed on an RTX A5000 GPU for half a day.
To reuse the weights on other base models, we simply compute the low-level features of the keyframes with the low-level feature encoder from MiVOS.
The corresponding overhead is already included in the reported times.

\addcontentsline{toc}{section}{Soft Motion-Vector Propagation Module}
\section*{Soft Motion Vector Propagation Module}

\textbf{Decoder $\mathcal{D}$:} The lightweight decoder in the soft propagation module as given in Equation 9 of the main paper is composed of two convolutional layers and three residual layers.  Specifically, it has
{\small
\begin{verbatim}
Conv2d(289,128)->Conv2d(128,64)->
ResBlock(64,64)->ResBlock(64,64)->
ResBlock(64,64)->Conv2d(64,11)->Upsampling.
\end{verbatim}}

{For all the layers, the kernel size is 3 with a padding of 1. We use BasicBlock of ResNet18 here as the ResBlock without the batch normalization. Final predictions are up-sampled by 4 with bi-linear interpolation.}

\addcontentsline{toc}{section}{Residual Correction Module}
\section*{Residual Correction Module}

\textbf{Feature matching}
$\tilde{\mathbf{S}}$ provides an indication of which areas in the propagated mask will require correction. For each pixel in $\tilde{\mathbf{S}}$ indexed by $a$ at frame $n$, we search in the temporally closest keyframe $k^*$ and match between $V_{n}$ and $V_{k^*}$. Specifically, we define $\mathbf{W}^{ak}$ as the affinity between the feature at pixel $a$ in $V_{n}$, \ie $V^a_{n}$, and all pixels in $V_{k^*}$. The corrected mask prediction at pixel $a$ is then obtained by $P^a_n = \mathbf{W}^{ak} P_{k^*}$. {Following STCN\cite{STCN}, we use an L2-similarity function to compute the affinity matrix, where $\mathbf{W}^{ak} \in \mathbb{R}^{1\times HW}$:}

\begin{equation}
     (\mathbf{W}^{ak})^b =\frac{f(V^a_{n}, V_{k^*}^{b})}{\textstyle\sum_m(f(V^a_{n},V_{k^*}^{m}))}.
\end{equation}\label{residual_patch}
\noindent $f$ is the L2 similarity function: 
\begin{equation}
f(V^a_{n}, V_{k^*}^{b})\!=\!exp(-||V^a_{n}\!-\!V_{k^*}^{b}||^2_2). 
\end{equation}\label{L2_sim}

\addcontentsline{toc}{section}{Timing Analysis}
\section*{Timing Analysis}

As analyzed in Section 5.4, we estimate the amortized per frame inference time via
\begin{equation}
T_{\text{base}}\!\cdot\! R +(T_{\text{propagation}}\!+\!T_{\text{correction}})\!\cdot\! (1\!-\!R),    
\end{equation}\label{amortized-time}
where $R$ denotes the ratio of keyframe.  $T_{\text{base}}$ denotes the inference times of the base segmentation model, while $T_{\text{propagation}}$ and $T_{\text{correction}}$ denote the time for motion-vector-based propagation, and time for residual correction, respectively. We measured the propagation and correction time on DAVIS17 with an RTX-2080Ti, and the sum ($T_{\text{propagation}}\!+\!T_{\text{correction}}$) is $12$ms.

\subsection*{Base model timing $T_{\text{base}}$} 
One of the key variable components in the timing is $T_{\text{base}}$.  Our reported values do not match the published FPS of the respective works for the base models.  When only keyframes are fed into the base model, the $T_{\text{base}}$ for fine-tuning-based models will be higher but shorter for memory-based models.  We elaborate on the details below. 

\textbf{Fine-tuning based model} contains two components for timing: online fine-tuning and segmentation. Applying the model to fewer frames, \ie~only on the keyframes, will reduce the segmentation time, but the online fine-tuning time remains the same. This leads to a higher $T_{\text{base}}$ than the published FPS, as online fine-tuning dominates and is less amortized over the various frames.  \cref{tab:timeanalysisfine-tune} shows $T_{\text{base}}$ for a fine-tuning based base model FRTM~\cite{FRTM}. Adding on our framework is less ideal, because the overall FPS largely depends on how much time the model spends on online fine-tuning, and reducing the number of frames processed by the base model does not provide much speed-up.  
\begin{table}[H]
\footnotesize
  \caption{FRTM~\cite{FRTM} base network timing analysis on DAVIS17}\label{tab:timeanalysisfine-tune}
  \centering
  \begin{tabular}{lc}
    \hline
    Frames used by base model FRTM~\cite{FRTM}  &$T_{\text{base}}$(ms)   \\
    \hline
    All frames in sequence  &  71.0    \\
    Only keyframes (37.2\% of all frames)    &  115.3     \\
    \hline
  \end{tabular}
\end{table}

\textbf{Memory-based models} also have two components for timing: memory reading/update and segmentation.  Using fewer keyframes can reduce both components. The total memorized frames are directly proportional to the total number of frames applied to the base model multiplied by the update frequency.  Keeping the same update frequency as the original base model results in a memory proportional to the percentage of keyframes. This is equivalent to increasing the interval between the stored frames in the memory for the original video sequence.

On DAVIS17, decreasing the update frequency reduces the keyframe accuracy. As shown in \cref{tab:timeanalysisDavis}, keeping the same update frequency for keyframes uses only 36\% of default memory, but results in a near 0.5 point drop on both $\mathcal{J}$ and $\mathcal{F}$ scores. Yet, updating with every keyframe would exceed the original frequency interval used during training.  This incompatibility 
causes the accuracy to drop. Additionally, it exceeds the original memory by 81\% and is nearly 46ms slower than the default memory setting.  In our reported result in Section 5.2 of the main paper, we make a trade-off between efficiency and accuracy and update every two keyframes.  This gives base times and a memory size approximately equal to the original base model.  

\begin{table*}[h]
\footnotesize
  \caption{Time and accuracy analysis on DAVIS17 on RTX 2080Ti (default encoder settings, 36.1\% keyframes). The star denotes the setting used for our main result in Section 5.2. The frame \% in memory is tabulated as a percentage of the original base model when applied to all frames of the sequence.}\label{tab:timeanalysisDavis}
  \centering
  \begin{tabular}{lcccccc}
    \hline
    Model & Update frequency & Frame \% in memory &  $T_{\text{base}}$(ms)& $\mathcal{J}$&$ \mathcal{F}$ & \\
    \hline
    MiVOS\cite{MiVOS} & every 5 frames & 100\% (by default) & 89.3 & 81.7 & 87.4 \\    
    MiVOS on keyframes & every keyframes  &181\% & 134.8 & 78.9 & 83.9\\
    MiVOS on keyframes & every 2 keyframes*  & 90\% & 87.9 & 79.7 & 84.6\\    MiVOS on keyframes & every 3 keyframes  & 60\% & 78.9 &  79.3&84.2 \\
    MiVOS on keyframes & every 5 keyframes  & 36\% & 73.8 & 79.1 & 84.1&\\
    \hline
  \end{tabular}
\end{table*}

\begin{table*}[h]
\footnotesize
  \caption{Time analysis for base model on the first 30 sequences of YouTube-VOS on RTX A5000 (default encoder preset, 27.2\% keyframes). We do not have access to the per-frame segmentation accuracies as evaluation scores and tallied privately over a test server.  Here, we report the combined score $\mathcal{G}$ averaged over all the frames for all sequences. The star denotes the setting used for our main result in Section 5.2. The frame \% in memory is tabulated as a percentage of the original base model when applied to all frames of the sequence.}  \label{tab:timeanalysisYT}

  \centering
  \begin{tabular}{lcccccc}
    \hline
    Model & Update frequency & Frames \% in memory &  $T_{\text{base}}$(ms)& $\mathcal{G}$ & \\
    \hline
    MiVOS\cite{MiVOS} & every 5 frames & 100\% (by default) &77.0 & 82.6 \\
    MiVOS on keyframes & every keyframe  & 135\% & 96.3 &  79.3\\
    MiVOS on keyframes & every 2 keyframe  & 68\% & 58.3 &  79.4\\
    MiVOS on keyframes & every 3 keyframe  & 45\% & 49.6 &  79.5\\
    MiVOS on keyframes & every 5 keyframes*  & 27\% & 47.0  & 79.3\\

    \hline  
  \end{tabular}
\end{table*}

In Table \cref{tab:timeanalysisYT}, we find that for YouTube-VOS, updating the memory every five keyframes achieves comparably accurate segmentations. As the original base model has a setting of updating every five frames over the entire sequence, this is the setting we choose for our results in Section 5.2 of the main paper. Updating every keyframe would result in 35\% more memory and 49ms slower at segmentation.

There is a difference in the memory update frequency between DAVIS and the YouTube-VOS dataset because 1)  YouTube-VOS has a higher frame rate than DAVIS (30FPS vs. 25FPS), and 2) YouTube-VOS is less dynamic in terms of object appearance. The I-/P-frames ratio between YouTube-VOS and DAVIS is 27.2\% and 36.1\%, respectively. Under the same encode setting, YouTube-VOS results in 10\% fewer I-/P-frames than DAVIS. Lower temporal density makes YouTube-VOS require less frequent memory updating.

\begin{table}[h]
\footnotesize
  \caption{Warping method towards target objects of different sizes on propagated frames on DAVIS17. 
  Each entry shows $\mathcal{J}\&\mathcal{F}$ scores of non-keyframes. 
  Keyframes are selected under the default encoder preset.}
  \label{tab:different_sizes_10000_30000}
  \centering
  \begin{tabular}{lccc}
    \hline
    &\multicolumn{3}{c}{DAVIS17}\\
    & small  & medium & large  \\
    \hline

    Optical Flow    &60.9&75.1&81.0\\
    
    MV to Flow      &60.0&73.0&77.8\\

    Ours  &72.7&89.7&91.9\\
    \hline
    Base model without propagation\cite{MiVOS} &76.8&91.9&92.3
                    \\

    \hline

  \end{tabular}
\end{table}

\addcontentsline{toc}{section}{Impact of Object Size}
\section*{Impact of Object Size}

To better understand the capabilities of mask propagation using motion vector warping, we further evaluate according to different sized objects.  We split the object masks into three categories based on their total pixels and report the segmentation accuracy in \cref{tab:different_sizes_10000_30000} 
according to object size.  Small objects have less than 10k pixels,~\ie~approximately $100\!\times\!100$ or smaller, medium objects are between 10k and 30k pixels,~\ie~between approximately $100\!\times\!100$ and $170\!\times\!170$ pixels, and large objects have more than 30k pixels,~\ie larger than $170\!\times\!170$ pixels.  This results in an approximately even split of 32\%, 35\%, 33\% for DAVIS17.  As expected, segmentation performance is proportional to the object size category, as smaller objects have less support in the scene and are therefore more challenging. 

Similar to the results in Table 2 of the main paper, we observe that our bi-directional and multi-hop motion vector propagation's accuracy exceeds "Optical Flow" warping and naive "MV to Flow"~\cite{CoViAR} by a large margin on DAVIS17. Compared to per-frame inference without any propagation, our proposed propagation and correction achieve excellent segmentation accuracies on medium and large objects. Interestingly, for the large objects, we are able to achieve comparable results with the original base model. This result validates the effectiveness of our motion vector-based soft propagation and residual-based correction.

{Our propagation and refinement scheme is less effective on small objects ($100\times100$). Small objects are challenging for many segmentation methods, and given the lightweight nature of our decoder, we observe a similar weakness in our system. 
We believe that the results can be improved if we choose to use deeper features and a more powerful decoder, but more computational cost will be applied.
}

\addcontentsline{toc}{section}{Qualitative Comparison}
\section*{Qualitative Comparison}

\begin{figure*}[h]
    \centering
    \includegraphics[width=\linewidth]{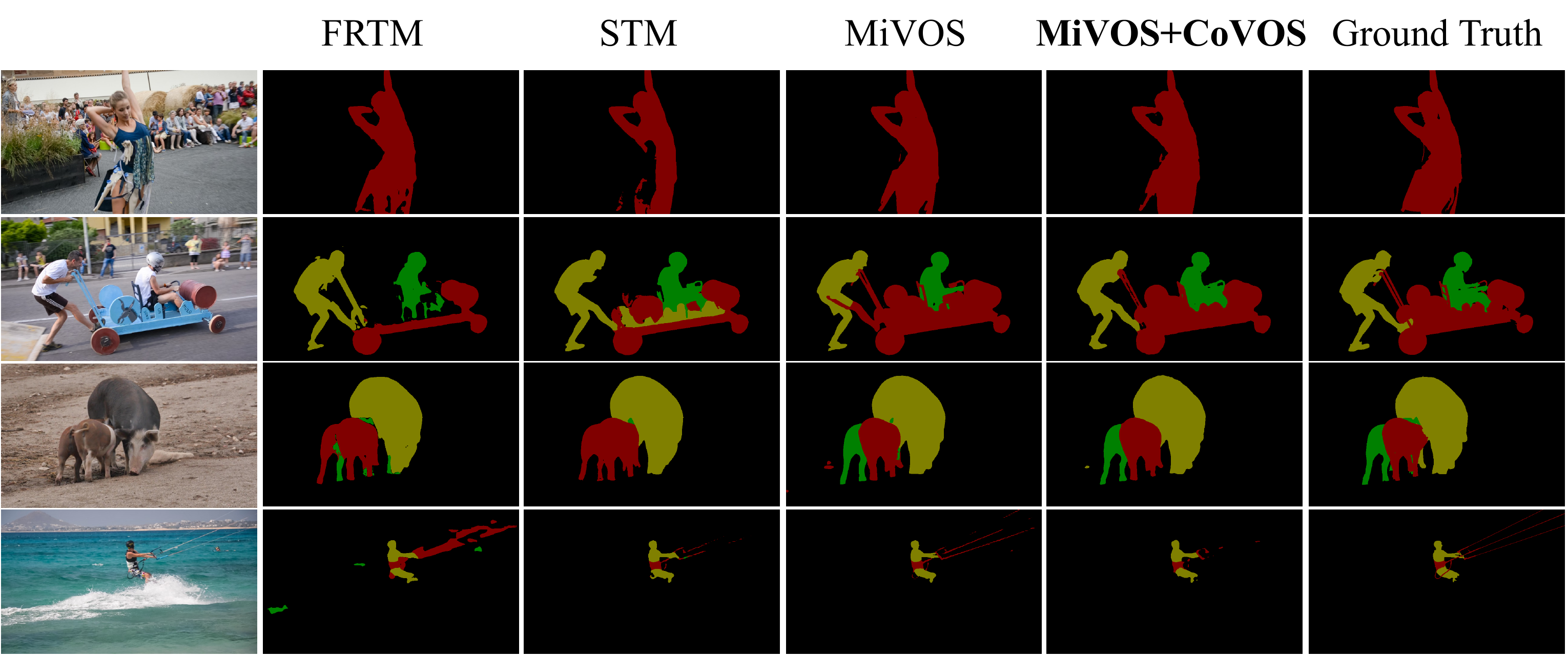}
    \caption{{Qualitative comparison with state-of-the-art methods. With MiVOS as the base model, we propagate most of the segmentation, while competing methods FRTM and STM exhibit several failures, e.g. fail to provide a precise mask (row 1: body of dancer, row 2: cart), distinguish similar objects (row 3: two piglets get merged) or provide more precise boundaries (row 4: kite handle). }} 
    \label{fig:qual}
\end{figure*}

\begin{figure*}[h]
    \centering
    \includegraphics[width=\linewidth]{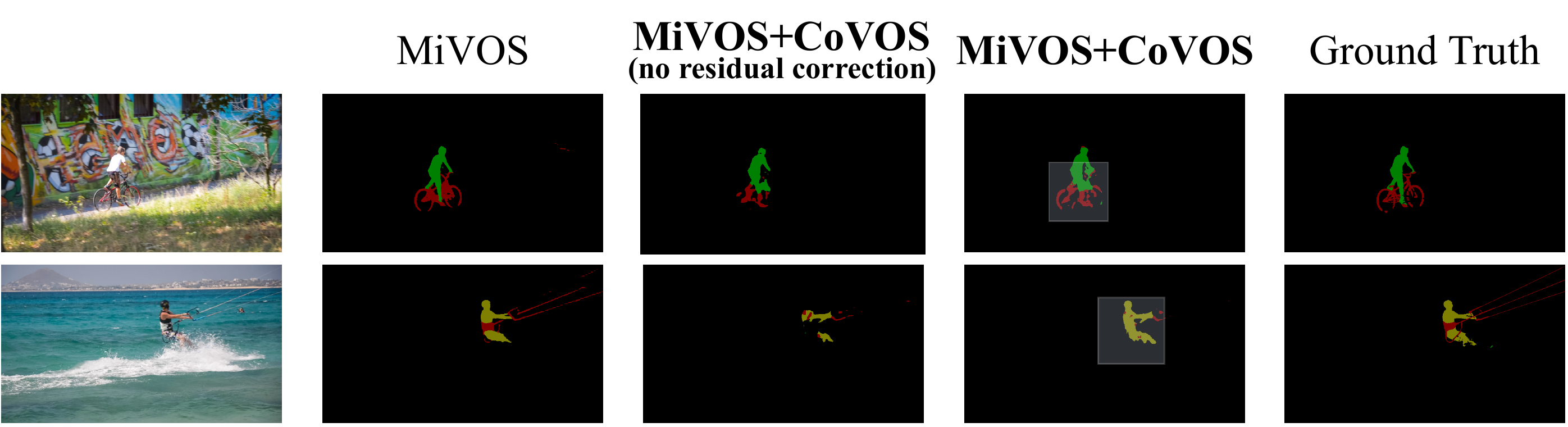}
    \caption{{Cases where motion vectors fail to correctly propagate the segmentation masks, but we recover the correct masks through residual correction. Label recovery through feature matching sometimes fails to distinguish similar regions as we only use very low-level features (second row).}}
    \label{fig:qual_res}
\end{figure*}

\begin{figure*}[h]
    \centering
    \includegraphics[width=\linewidth]{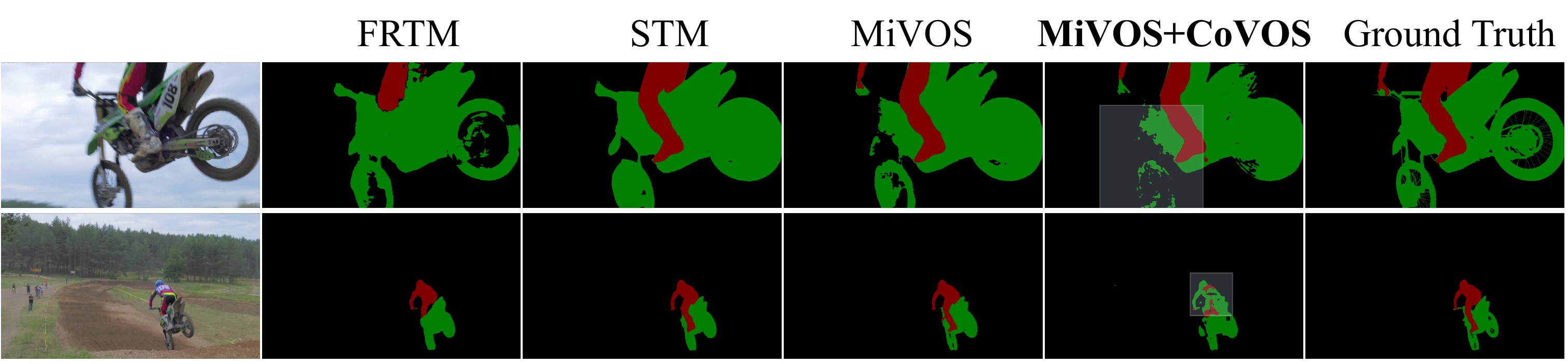}
    \caption{{Our method struggles in cases of abrupt motions, which lead to inaccurate motion vectors (see white rectangle highlights).}}
    \label{fig:qual_neg}
\end{figure*}

{
\cref{fig:qual} highlights success cases compared with other state-of-the-art methods and the base model, where our method can provide a full mask and recover most of the boundary detailing compared to the base model.

{\cref{fig:qual_res} shows some sample cases of our residual correction module. In the ~\textit{bmx-trees} sequence, we are able to recover the bike from the residual. Similarly, in the ~\textit{kite-surf} sequence, where most of the human body is missing in the propagation, we successfully recover it from the residual. However, we fail to estimate the correct label of the band due to its similar appearance to the human body.}

\cref{fig:qual_neg} shows 
some challenging cases for our methods. In the~\textit{motocross-jump} sequence shown in the first and second row, the motion vector fails to capture the fast scene change, resulting in inaccurate segmentations as highlighted with white rectangles.
}

We show more qualitative examples on the DAVIS dataset in Figure 4 and 5.

\begin{figure*}[h]
    \centering
    \includegraphics[width=0.9\linewidth]{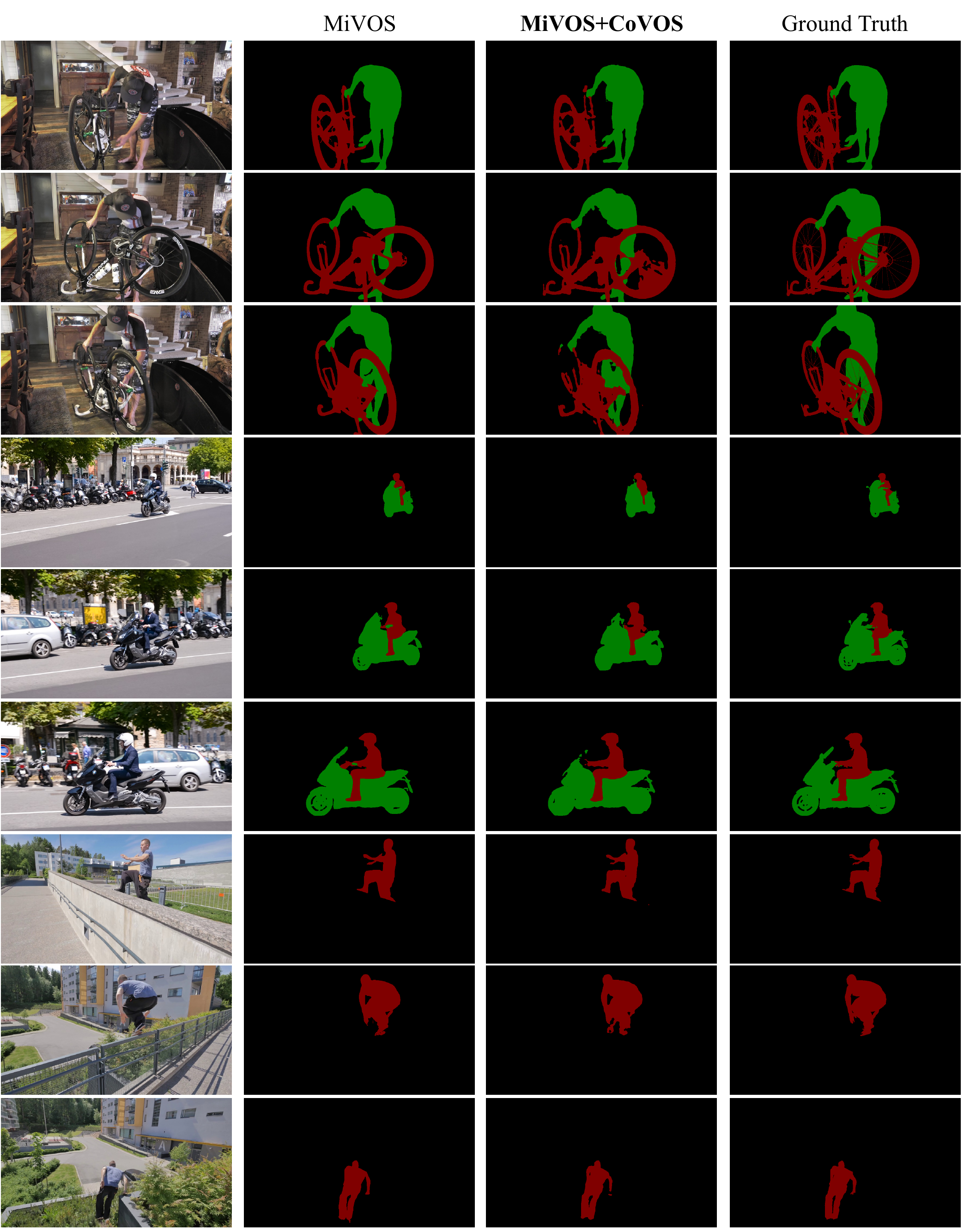}
    \caption{Qualitative comparison on DAVIS17.}
    \label{fig:more_qual1}
\end{figure*}

\begin{figure*}[h]
    \centering
    \includegraphics[width=0.9\linewidth]{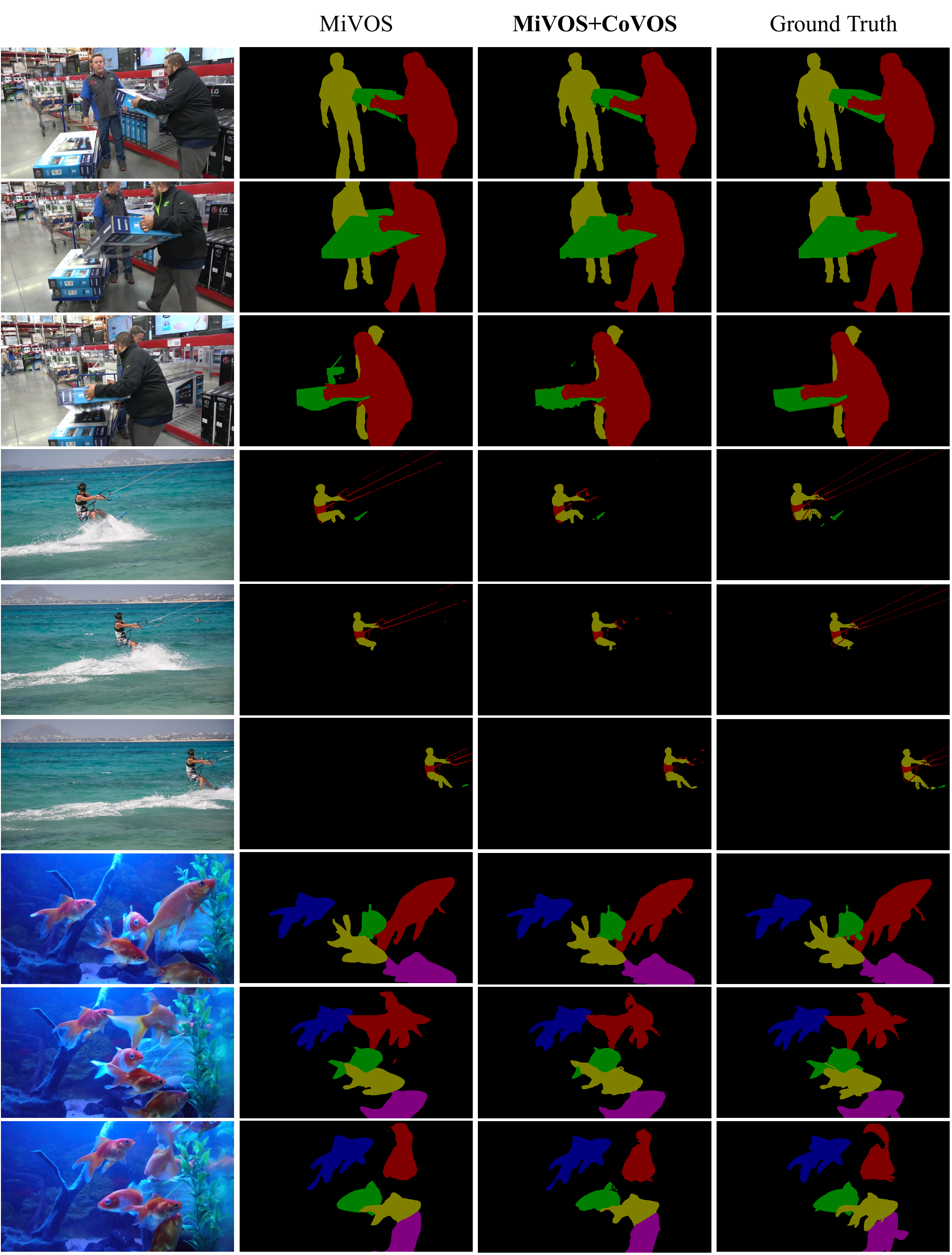}
    \caption{Qualitative comparison on DAVIS17.}
    \label{fig:more_qual2}
\end{figure*}

\addcontentsline{toc}{section}{Licence}
\section*{Licence}
The annotations in DAVIS belong to the organizers of the challenge and are licensed under the \href{https://github.com/fperazzi/davis-2017/blob/master/LICENSE}{BSD License}. The annotations in YouTube-VOS belong to the organizers of the challenge and are licensed under a \href{https://creativecommons.org/licenses/by/4.0/}{Creative Commons Attribution 4.0 License}. Code and pretrained weights for \href{https://github.com/hkchengrex/MiVOS/blob/main/LICENSE}{MiVOS}, \href{https://github.com/hkchengrex/STCN/blob/main/LICENSE}{STCN} and \href{https://github.com/andr345/frtm-vos/blob/master/LICENSE.txt}{FRTM-VOS} are under GNU General Public License v3.0.

{\small
\bibliographystyle{ieee_fullname}
\bibliography{reference}
}
\clearpage